\documentclass{article}

\usepackage{PRIMEarxiv}

\usepackage{amssymb}
\usepackage[utf8]{inputenc}
\usepackage{mathptmx}
\usepackage{graphicx}
\usepackage{hyperref}
\usepackage{subcaption}
\usepackage{verbatim}
\usepackage{mathrsfs}
\usepackage{caption}
\usepackage{tabularx}
\usepackage{placeins} 
\usepackage{stfloats} 
\usepackage{natbib}
\usepackage{multirow}
\usepackage{amsmath}
\usepackage{algorithm2e}

\usepackage{array}
\usepackage{longtable}

\usepackage[T1]{fontenc}    
\usepackage{url}            
\usepackage{booktabs}       
\usepackage{amsfonts}       
\usepackage{nicefrac}       
\usepackage{microtype}      
\usepackage{lipsum}
\usepackage{fancyhdr}       
\usepackage{graphicx}       
\graphicspath{{media/}}

\title{From Text to Graph: Leveraging Graph Neural Networks for Enhanced Explainability in NLP}

\author{
  Fabio Yáñez-Romero \\
  University Institute for Computer Research \\
  University of Alicante \\
  Alicante, Spain \\
  \texttt{fabio.yanez@ua.es} \\
  \AND
  Andrés Montoyo, Armando Suárez, Yoan Gutiérrez \\
  Department of Computing and Information Systems \\
  University of Alicante \\
  Alicante, Spain \\
  \texttt{\{montoyo, armando, ygutierrez\}@dlsi.ua.es} \\
  \AND
  Ruslan Mitkov \\
  Lancaster University \\
  Lancaster, United Kingdom \\
  \texttt{r.mitkov@lancaster.ac.uk} \\
}

\begin{document}
\maketitle

\begin{abstract}
Researchers have relegated natural language processing tasks to Transformer-type models, particularly generative models, because these models exhibit high versatility when performing generation and classification tasks. As the size of these models increases, they achieve outstanding results. Given their widespread use, many explainability techniques are developed based on these models. However, this process becomes computationally expensive due to the large size of the models. Additionally, transformers interpret input information through tokens that fragment input words into sequences lacking inherent semantic meaning, complicating the explanation of the model from the very beginning. This study proposes a novel methodology to achieve explainability in natural language processing tasks by automatically converting sentences into graphs and maintaining semantics through nodes and relations that express fundamental linguistic concepts. It also allows the subsequent exploitation of this knowledge in subsequent tasks, making it possible to obtain trends and understand how the model associates the different elements inside the text with the explained task. The experiments delivered promising results in determining the most critical components within the text structure for a given classification.

\end{abstract}


\section{Introduction}\label{Introduction}

Large language models (LLMs) have proven useful for numerous tasks in natural language processing, from low-level tasks related to natural language understanding, such as dependency tree creation \cite{hromei2023end}, semantic role labelling (SRL) \cite{shi2019simplebertmodelsrelation}, and textual entailment (TE) \cite{Garc_a_Silva_2023}, to high-level tasks, such as text generation variants like continuing a text \cite{becker2024textgenerationsystematicliterature}, generating summaries \cite{zhang2024systematicsurveytextsummarization} or answering questions \cite{nachmani2024spokenquestionansweringspeech}.

In most cases, LLMs are based on transformer architectures, which are born from the use of attention mechanisms applied to natural language processing. These architectures can consider a greater sequence context than recurrent neural networks (RNNs) and Long Short-Term Memory networks (LSTMs) and improve the results on these networks.
 
The widespread acceptance of transformer models in NLP means that explainability techniques aimed at discovering why the model behaves in one way or another focus on transformer architecture. However, this poses a serious problem for the following reasons:

\begin{enumerate}
    \item  Transformer models vectorise the input text through a tokeniser. This tokeniser fragments the text into sequences that may lack semantic meaning for humans. This is a problem if looking for explainability that is interpretable for humans since the way humans carry out NLP tasks considers the semantics of the elements of the sentence.

    \item The internal architecture of transformer models does not promote explainability; in each layer of the transformer model, there is an attention mechanism, which makes it possible to take into account the entire input sequence, and there are also multilayer perceptrons (MLPs) which store the information captured by the attention mechanism. Although the attention mechanism allows for the focus on specific input sequences, the model implicitly learns the semantic rules of the language by processing large amounts of text that present similar patterns. These learned patterns can cause the model to hallucinate since the semantic rules learned are not explicit. 
\end{enumerate}

In addition, if we consider that ever-larger LMs tend to be developed because they improve their performance with the model's scalability and training data, applying sampling techniques to determine explainability in competitive models becomes exceptionally costly.

Other models, such as graph neural networks (GNNs), allow semantic relationships to be explicitly preserved during the learning stage, allowing for greater traceability.  The most commonly used implementations of these networks perform a convolution around a central node, similar to the pixels used in convolutional neural networks (CNNs), with the particularity that in GNNs, the number of neighbours for performing the convolution is variable. The greater traceability of these networks has sparked great interest in the scientific community in developing different explainability modules. Likewise, these networks have become popular for tasks that do not initially involve node or graph classification, such as natural language processing tasks, in which the initial text sequence is translated into a graph.

We hypothesise that explainability is achievable in natural language processing by focusing on structured formats such as graphs instead of transformer models. Thus, gaining greater traceability and ensuring that explainability is easy to interpret enables us to obtain information that allows us to better understand the behaviour of the models and use the given results in subsequent tasks. Achieving this explainability without seeking it within the architecture of a transformer model will allow explainability techniques to be used with few resources. Associating them with concepts with semantic meaning will ensure no loss of traceability concerning the initial data.

This study presents a novel methodology for natural language processing tasks that takes advantage of the structure of graphs, specially designed for natural language processing, to gain traceability and explainability in the results obtained. Thus, it can determine the most critical nodes and relations in the decisions made for semantic classification tasks, such as topic classification, sentiment analysis, or emotion detection. In this way, the implicit knowledge of the different sentences of the language models is represented explicitly through constituency relations. Increased traceability and explainability can help perform model analytics, diagnose the model itself, and have a more reliable method for automatically populating knowledge bases \cite{zhong2023comprehensivesurveyautomaticknowledge}.

The paper is structured as follows: To start with, research related to this publication is discussed (Section \ref{related_work}). This section covers the recent advances in explainability for language models, neural networks, graphs, and the automatic conversion of text to structured data. Following this, the proposed methodology is detailed (Section \ref{methodology}), outlining all the steps of the process and its underlying logic. After describing the methodology, the focus shifts to the experiments conducted (Section \ref{experiments}), with a presentation of the results. A discussion of the results follows (Section \ref{discussion}), leading to the conclusions of the publication (Section \ref{conclussion}) and suggestions for future work based on the findings (Section \ref{future_work}).

\section{Related Work} \label{related_work}


To understand the proposed methodology, it is essential to mention how explainability is handled in language models and its problems (Subsection \ref{explainability_lm}), how the text can be translated into a structured format that retains all its prior knowledge (Subsection \ref{text_trees}) and the main advantages of using explainability in graphs (Subsection \ref{explainability_graphs}).

\subsection{Explainability in Language Models} \label{explainability_lm}
Explainability in language models is a growing area of research that aims to mitigate problems inherent in large generative models, such as hallucination, or bias and toxic language, control the model's behaviour, and diagnose issues in the model. The high cost of training language models also motivates the explainability of pre-trained models, since trying to understand how the model works can be less expensive than retraining from scratch, especially with the larger models that tend to be used for text generation tasks \cite{touvron2023llamaopenefficientfoundation}.

The primary methodologies used to explain language models are divided in local and global analysis methods according to \cite{luo2024understandingutilizationsurveyexplainability}. Local methods focus on analysing the internal structure of language models, while global methods interpret the model by sampling it in different ways. 


In the case of local methods, some methods determine the relevance of each token in the prediction of the model, either by altering the data input (\cite{modarressi-etal-2023-decompx}, the study of gradients associated with the different tokens \cite{kindermans2017unreliabilitysaliencymethods}, or the representation of the various tokens in their respective vectors along the model \cite{modarressi-etal-2022-globenc}. Other local methods seek to study the behaviour of the transformer models in a granular way, distinguishing at this point those that focus on the analysis of the attentional mechanism \cite{kobayashi-etal-2020-attention} and on the multilayer perceptron (MLP) \cite{geva-etal-2022-transformer}.

Global methods include model probing, which focuses on probing the knowledge acquired by the model according to its initial training set \cite{peng2022copenprobingconceptualknowledge} and knowledge representations with classifiers \cite{li2024inferencetimeinterventionelicitingtruthful}. Other kinds of global methods are mechanistic interpretability models, which focus on the connections between different parts of the model globally, such as aligning model representations with ground truth \cite{chughtai2023toymodeluniversalityreverse}, studying how neuron activation affects model output globally \cite{meng2023masseditingmemorytransformer} or locating areas of the model associated with specific knowledge \cite{belrose2023elicitinglatentpredictionstransformers}. 

Although different methods have emerged in recent years to explain the behaviour of language models, these methods are based on unstructured knowledge, such as text. In the case of tokens used in language models, this knowledge can be even less structured, dividing words into different tokens that have no inherent semantic meaning.

Tree structures, which can be interpreted as graphs, can be a starting point for translating unstructured to structured knowledge, retaining all the information from the original text. They capture the information of each sentence's words and the relationships between the elements of the text. In this context, dependency and constituency trees represent sentences as graphs that maintain all the information.

\subsection{Dependency and Constituency trees in NLP} \label{text_trees}

Unstructured information in texts can be represented in a structured way using linguistic rules that relate different words to each other or generate subgroups within each sentence. In this case, three different types of trees can be distinguished in sentence-level natural language processing:

\begin{itemize}
    \item[] · Syntactic dependency trees represent the different words of the sentence according to syntactic relations between them, with the sentence's main verb being the tree's origin.
    \item[] · Semantic dependency trees represent the semantic relations between a sentence's different words, distinguishing between one or several predicates formed by the main verbs of the sentence and arguments.
    \item[] · Constituency trees represent the sentence structure hierarchically with phrase rules, from the main components to the most secondary ones.
\end{itemize}

Dependency and constituency tree generation models became popular with Bidirectional Long Short-Term Memory (BiLSTM) \cite{e22080870}, first enhanced with Biaffine attention \cite{dozat2017deep} and later by Transformer-like architectures instead of BiLSTM \cite{sun2020selfexplainingstructuresimprovenlp}. Beyond that, the Automated Concatenation of Embeddings \cite{wang2021enhanceduniversaldependencyparsing} has been used to find the best combination of embeddings in these tasks, as well as reinforcement learning techniques on the above-mentioned methodologies \cite{10.1145/3616855.3635775}. Heterogeneous Syntax Fuser (HeSyFu), introduced by \cite{fei2021better}, combines syntactic dependency trees with constituency trees using graph convolutional networks (GCNs) \cite{kipf2017semisupervised} and Tree-LSTM \cite{tai2015improvedsemanticrepresentationstreestructured} to perform Semantic Role Labelling (SRL).

It is clear that obtaining dependency and constituency trees has a long way to go until it almost surpasses, in many cases, the efficiency of expert annotators in the field. This motivates the use of automatic conversion from sentences to graphs to obtain more explainable structures thanks to the inherent traceability of graphs and the significant development of explainability techniques in GNNs.

\subsection{Explainability for Graph Neural Networks} \label{explainability_graphs}

One of the advantages of graph neural networks over language models or other deep learning architectures is the explainability of graphs through different methods. As an architecture, graphs are considered easier to explain than text due to their inherent structure, with nodes representing specific elements and their existing relationships. As long as the nodes and relationships represent concepts humans understand, we can start from a structure that enables interpretability. Moreover, the operations performed by the most widely used graph neural networks tend to apply convolutions by assimilating the information of neighbouring nodes in each of the central nodes considered, which implies greater traceability than conventional neural networks, which are fully connected. In addition, GNNs are usually lighter than the LMs used for the same purpose, which leads to a lower computational cost when achieving explanations using such models.

According to surveys conducted on explainability in spatial graph neural networks \cite{kakkad2023surveyexplainabilitygraphneural}, techniques can be classed as \textquotedblleft factual\textquotedblright{} if they try to determine the features that cause the result obtained or \textquotedblleft counterfactual\textquotedblright{} if focusing on patterns from the input data that can cause a change in the results.

Within the factual methods, we can consider post-hoc methods, which are characterised by not modifying the original structure of the GNN to explain it and using the parameters of the frozen model to develop the explanation. Various post-hoc methods, such as decomposition-based methods, treat the prediction as a score to be decomposed from the end of the GNN to the beginning \cite{8954227}. Gradient-based methods pay attention to the change that occurs in the gradient with changes in the input data by determining which features of the input data are the most sensitive \cite{Selvaraju_2019}. Surrogate methods attempt to approximate the relationship between the input data and the results obtained with interpretable lightweight models coupled with the frozen trained GNN \cite{vu2020pgmexplainerprobabilisticgraphicalmodel}. The perturbation-based methods focus on determining which elements of the initial graph are the most determinant for the prediction obtained, removing different elements from the original graph, thus giving the subgraph that assures the predicted result \cite{ying2019gnnexplainergeneratingexplanationsgraph}. On the other hand, generation-based models use graph generators or generative models to determine which input patterns cause the result obtained \cite{NEURIPS2021_be26abe7}. 

Another alternative within the factual methods is the self-interpretable methods that modify the original GNN to explain. These methods use structural constraints on the model \cite{wu2022discoveringinvariantrationalesgraph} or the input information \cite{miao2022interpretablegeneralizablegraphlearning} to obtain a graph that provides both the model's final output and the model's explainability.

Counterfactual methods can be classified into perturbation-based methods, the most classical of which modify the edges of the graph to study their effect from changes in the adjacency matrix \cite{lucic2022cfgnnexplainercounterfactualexplanationsgraph}. Neural methods generate counterfactual graphs with neural networks to perturb the original graph minimally, optimising the minimum change in the neural network used (\cite{bajaj2022robustcounterfactualexplanationsgraph}). Finally, search-based methods try to find similar graph elements within the set of employed graphs that give different results \cite{numeroso2021meggeneratingmolecularcounterfactual}.

\subsection{Relation Between Graphs and Language Models}

Considering graphs in other formats, \cite{ye-etal-2024-language} presents different graph construction methodologies to perform natural language processing tasks with reasoning over the generated graphs. They translate the graph architecture into a prompt and carry out the proposed task in generation language models. The methodology used by \cite{fatemi2023talk} involves the automatic generation of graphs with different algorithms, which are not linked to language rules but are algorithms that tend to create graphs of different densities by determining the number of edges and their distribution. These works show that language models can understand the data structure of a graph, both to process it as input data and generate it through text.

From a Geometric Deep Learning point of view \cite{bronstein2021geometric}, transformer models represent an internal vector space similar to a fully connected graph, as each token pays attention to the other tokens in the input sequence. Therefore, we consider our approach a constraint of the original space based on explicitly stated language rules and a space with less knowledge that can avoid confusion within the model by using much less data during training.

The explainability of language models is a significant topic within the scientific community. Still, the traceability of this explainability is linked to the tokens representing the text, which can obscure semantic meaning and complicate the application of obtained knowledge in subsequent tasks. 

To address this issue, utilising structured knowledge that converts graphs to text automatically is proposed. This method enhances language processing tasks in environments with greater traceability and explainability, leveraging the various explainability modules available for graphs. Additionally, this approach can lead to substantial computational savings, given the size differences between language models and graph neural networks.

\section{Methodology} \label{methodology}

\begin{figure}[]
    \centering
    \includegraphics[width=0.8\linewidth]{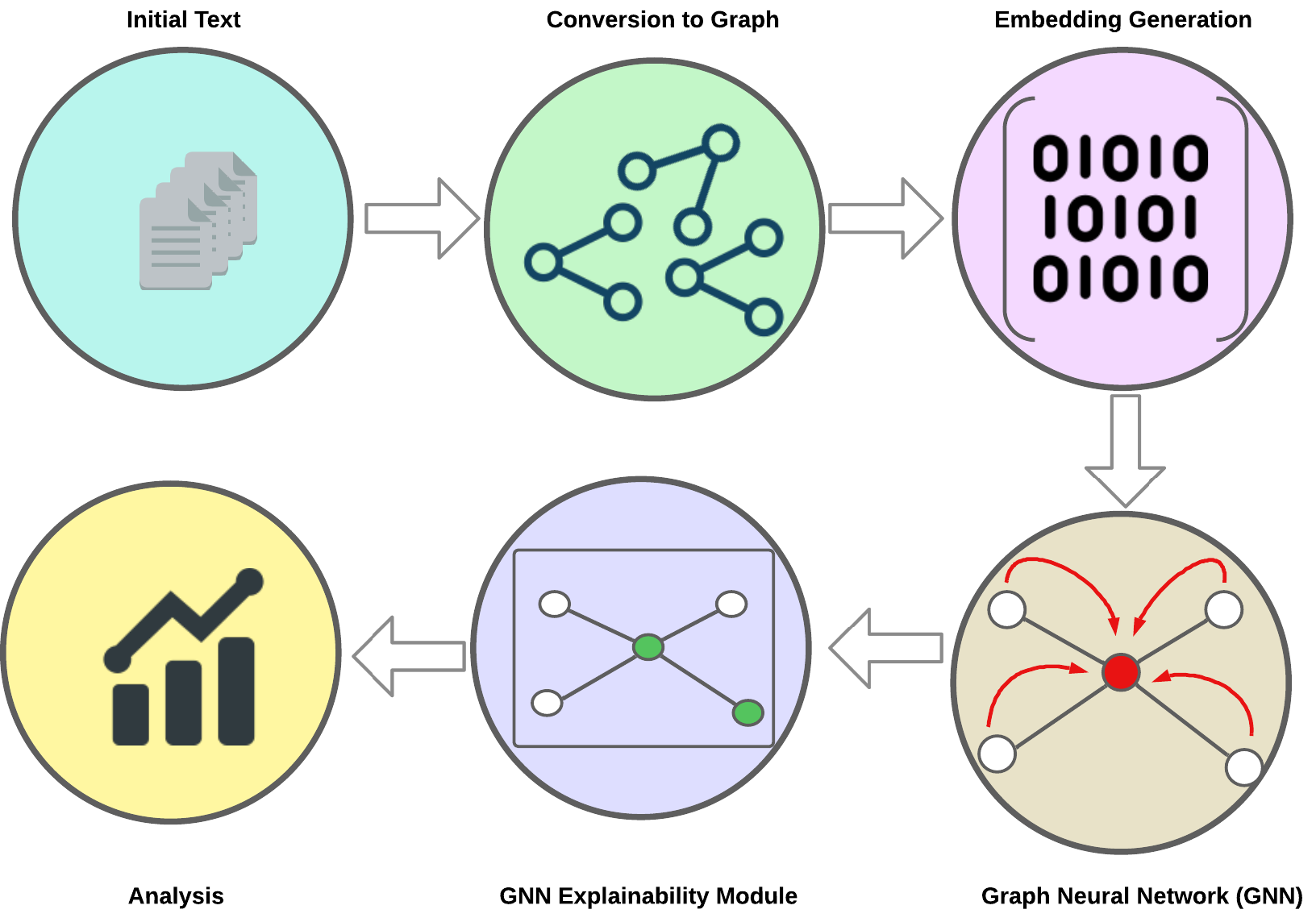}
    \caption{The proposed methodology covers converting the original texts into graphs and initialising their tensors automatically. Subsequently, explainability can be covered using neural network graphs, and relevant analyses of these results can be carried out.}
    \label{fig_methodology}
\end{figure}

\begin{figure}
    \centering
    \rotatebox{90}{%
        \begin{minipage}{\textheight} 
            \centering
            \includegraphics[width=1.1\textheight]{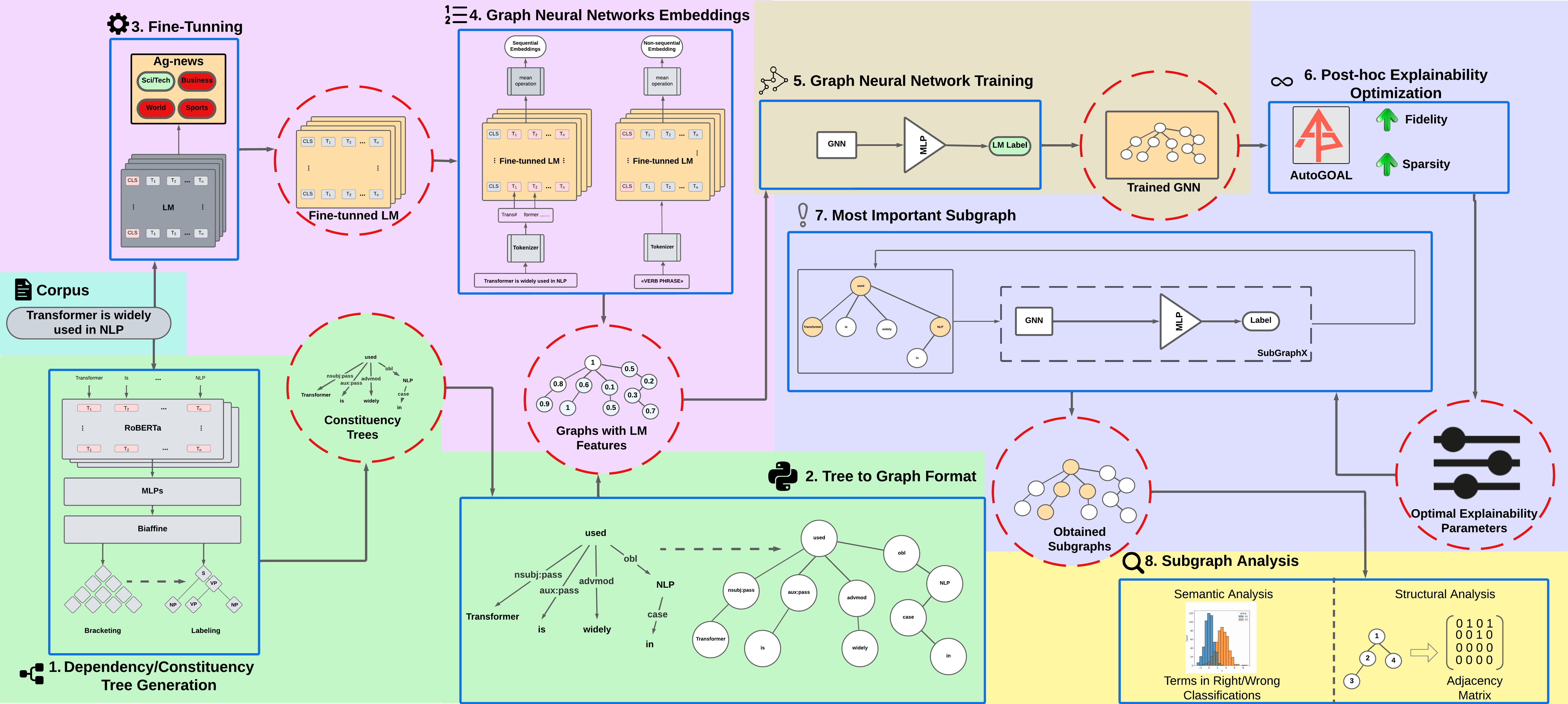}
            \caption{Our approach achieves explainability in a human-interpretable form through graph structures. In this case, an encoder-only language model explains the results obtained in topic classification using Ag-news and sentiment analysis using Sst-2. We start with language models fine-tuned in this task to generate the embeddings of the graphs.}
            \label{fig_architecture}
        \end{minipage}
    }
\end{figure}

Our proposal focuses on maintaining sentence structures that have semantic meaning, in our case words, as opposed to the traditional approach of Transformers models that treat text sequences by splitting words into sequences that lose this semantic meaning. The proposed methodology is illustrated in Figure \ref{fig_methodology}, and its explanation and current implementation are discussed in detail in the following subsections and shown in Figure \ref{fig_architecture}; the same colour in both the methodology and implementation represent each stage. First, the rationale for the text-to-graph conversion is provided, as well as the category in which we can classify this method of explicability (Subsection \ref{llm_distillation}). Second, the text is converted into a tree (Subsection \ref{data_preprocessing}), keeping all the information of the original structure, and each tree is translated into a graph (Subsection \ref{data_conversion}). Then, the language model is fine-tuned for the task for which we want to achieve explainability (Subsection \ref{fine-tuning}), using this model to generate embeddings of the different elements of the previously obtained graphs (Subsection \ref{embeddings}). The embeddings and the graph structures are employed to train a graph neural network to replicate the language model (Subsection \ref{gnn}). Then, an explainability module is applied, optimising the model's hyperparameters to obtain high certainty in the results (Subsection \ref{optimization}). Analysing these graphs according to the explainability module can give us information on the importance of each term associated with the covered task (Subsection \ref{analysis}). The code required to replicate the experiment is available in the following GitHub\footnote{\href{https://anonymous.4open.science/r/Graphs_from_text-3C7A/}{Repository for Experiment Reproducibility}.} repository.

To the best of our knowledge, no sentence structure research has been done to achieve natural language processing explainability, and trim work has been done to distil language model knowledge into GNNs, as can be seen in \cite{mavromatis2023traingnnteachergraphaware}, \cite{li2024enhancinggraphneuralnetworks} or \cite{hu2024largelanguagemodelmeets}. 

\subsection{NLP Explainability through Distillation in GNN} \label{llm_distillation}

Text is considered a source of unstructured knowledge, as the knowledge it contains is implicit, i.e. it is not directly stated. In this case, there is a set of syntactic and semantic rules, constituency relations and, at a higher level of abstraction, coreference relations between entities, the communicative intention of a document (discourse) or the main topic of a document.

The knowledge inside a text is of interest in natural language processing tasks such as information extraction (IE), which is based on two lower-level tasks: identifying triplets of relations between entities and detecting coreferential links of the entities appearing in the document. However, these types of tasks are usually performed at the document level, and they are not feasible if the text is too shot, as in sentences.

In NLP, many tasks are successfully covered at the sentence level, such as obtaining syntactic, semantic, and constituency trees. These tasks are fundamental in other high-level natural language tasks, such as natural language understanding (NLU), Word Sense Disambiguation (WSD), Semantic Role Labelling (SRL), Machine Translation, Text Summarisation, Question Answering, Sentiment Analysis, etc. 

Given the competitive results in translating text into constituency trees, as discussed in Section \ref{related_work}, an automatic approach has been selected for processing sentences in text classification tasks. This recasts the problem as graph classification, allowing access to all resources related to this data architecture and having a lighter and more interpretable model than the LLM.

In this case, the compatible explainability methods from the language model point of view are global according to the survey on explainability on Language Models by \cite{luo2024understandingutilizationsurveyexplainability} since the extraction of embeddings and the use of the classifications obtained by the language model are ways of probing this model.
Moreover, the trained model (GNN) is more straightforward to interpret than the language model. This falls into the classification of surrogate explainability models, as we are trying to replicate the behaviour of the language model.
Additionally, we can consider this methodology a distillate of the original model since the GNN is trained using knowledge from the language model.
Finally, the real explanation of the model is carried out with post-hoc explainability modules on the obtained GNNs.

\subsection{Constituency Tree Generation} \label{data_preprocessing}

Performing a natural language processing task with GNNs requires generating graphs based on specific criteria from text data.

In our case, we generated constituency trees from pre-trained and fine-tuned language models for this task. RoBERTa \cite{liu2019robertarobustlyoptimizedbert} has been used and fine-tuned in the Structured Prediction of Morphology and Syntax (SPMRL) dataset \cite{tsarfaty2020spmrlnmrldidlearn} for this task as indicated in Figure \ref{fig_architecture}: Constituency Tree Generation. This way, the entire dataset is transformed from unstructured information, such as text, to structured information in tree format.

\subsection{Tree to Graph Conversion} \label{data_conversion}
The annotations the model returns provide us with the trees in the CoNLL-2009 format \cite{CoNLL-2009-ST}. From the results in this format, we have obtained the lists of the different words in each sentence, the relations obtained for each case, and the hierarchy of the tree, generating their respective directed graphs in the process with specific scripts as illustrated in Figure \ref{fig_architecture}: Tree to Graph Format.

\subsection{Language Model Fine-tuning} \label{fine-tuning}
The next step is to fine-tune the language model to the specific task we want to achieve explainability. In this way, the model's parameters are adjusted to perform the task correctly, and thus, the embeddings of this model can be used in subsequent tasks; these embeddings will be used as features of the GNN trained to replicate the Language model, as well as the classifications obtained from the Language Model on that specific task. In principle, the ideal models to perform this task are those with a CLS token since this token tends to store a representation of the entire input sequence. This token is found in models such as BERT \cite{devlin2019bert}, ALBERT \cite{lan2020albertlitebertselfsupervised}, Electra \cite{clark2020electrapretrainingtextencoders} or DeBERTa \cite{he2021debertadecodingenhancedbertdisentangled}. The fine-tuning step is illustrated in Figure \ref{fig_architecture}: Fine-tuning.

\subsection{Graph Neural Network Embeddings} \label{embeddings}

In graph theory, a graph is defined as a space formed by a set of elements called nodes \textit{V}, which can be related to each other with specific relations \textit{E}, so each graph is defined by a set of nodes and relations $G = (V, E)$. In our starting graphs, we have different entities, one for each word of the original sentence, which we have called word entities ($V_w$) and other entities representing special unique elements of the constituency trees ($V_s$) as \textit{Nominal Phrase} or \textit{Verbal Phrase} for instance. The set of different nodes our graphs consider is represented in Equation \ref{eq:nodes}.

\begin{equation}
\begin{aligned}
    V_w &= \{v_{w_{1}}, v_{w_{2}}, \dots , v_{w_{n}}\} \\
    V_s &= \{v_{s_{1}}, v_{s_{2}}, \dots , v_{s_{n}}\} \\
    V &= (V_w, V_s)
\end{aligned}
\label{eq:nodes}
\end{equation}




In the case of relations obtained by constituency trees $(R)$, we can consider only one type of relation without losing information since this information is fully represented by the graph hierarchy with its directed relations and the types of nodes considered.

The graphs obtained through constituency parsing models contain graph information in text format, which is unsuitable for GNNs. Therefore, the next step is to convert the nodes and relations of these graphs into tensors that can be processed in GNNs. In our case, we have chosen to use the fine-tuned language model to generate the embeddings of each graph. The method used to obtain the embedding of each node changes depending on its nature, according to the classification we have used. It is illustrated in Figure \ref{fig_architecture}: Graph Neural Networks Embeddings. 


In the case of $v_w$, each embedding is obtained through the tokens that form the complete word in the language model by passing the complete sentence in which that word appears as input to the language model, obtaining the last hidden states of those tokens. An aggregation is then performed by reducing all the tensors of that word to a single one, using the average of the tensors obtained. This method allows us to take advantage of the positional encoding used by the language models.

For $v_s$, a specific name is assigned to each one, as shown in \ref{tables} Table \ref{tbl:constituents}. The embeddings are obtained from the hidden state of the token [CLS], which tends to represent the whole input sequence. In this case, it is unnecessary to use positional encoding since we consider these nodes to be non-sequential elements, as they do not represent words.

\subsection{Graph Neural Network Training} \label{gnn}
Once we have all the necessary features to train the GNN, the training is performed with the given classifications of the fine-tuned language model, replicating its answers and evaluating the GNN with the same metrics as the LM, as illustrated in Figure \ref{fig_architecture}: Graph Neural Network Training. 

\subsection{Post-Hoc Explainability Optimisation}  \label{optimization}

Among the frameworks developed to explain graphs, we can find different post-hoc methods. These methods seek to explain the results obtained in the model once it has already been trained, as opposed to other methods that try to cover explainability from an early stage in the development of the model.
We have focused on the subgroup of post-hoc methods known as perturbation methods. These methods are characterised by trying to modify the data input to the trained model to determine which structures of the initial graph give the target result.
Considering the perturbation methods, we focus on SubgraphX \cite{yuan2021explainabilitygraphneuralnetworks}, which uses search techniques such as Monte Carlo Tree Search (MCTS) to determine which connected subgraph maintains the original classification given by the GNN. This method is much more interpretable since the elements of the graph that provide the explainability are connected, thus avoiding the generation of isolated subgraphs where specific nodes are unrelated as done in previous post-hoc explainability methods \cite{ying2019gnnexplainergeneratingexplanationsgraph}, \cite{vu2020pgmexplainerprobabilisticgraphicalmodel}, \cite{schlichtkrull2022interpretinggraphneuralnetworks}.

Although SubgraphX allows us to obtain the most critical subgraphs for each example, the result may give little information if the appropriate hyperparameters are not used for each situation. A poor hyperparameter selection can result in some of the subgraphs obtained being too large compared to the original graph, distilling too little information from this graph. Given that graphs can vary in size and connectivity within the same dataset, using a single set of hyperparameters for the whole dataset is counterproductive since, in some cases, we can obtain subgraphs that are too large compared to the original graph as the total number of nodes obtained in the subgraph is one of the hyperparameters, in the same way too few MCTS iterations can be performed on large graphs, the hyperparameters to be optimised in SubgraphX are illustrated in Table \ref{table:hyperparameters}. These hyperparameters have been optimised using AutoGOAL \cite{estevez2020general} through several experiments. Since the hyperparameters to be optimised are not attached to a usual training but are carried out in a specific framework such as SubgraphX, AutoGOAL is especially useful after defining the metrics we want to maximise since it is a general-purpose optimisation framework. Moreover, as demonstrated in previous research, autoGOAL can carry out multi-objective optimisation, one feature that optimisation frameworks do not usually incorporate \cite{estevanell2024balancing}.

\begin{table}[]
\resizebox{\textwidth}{!}{%
\begin{tabular}{c|c|c|c}
\hline
\textbf{Hyperparameters} & \textbf{Definition}                                                                                           & \textbf{Type} & \textbf{Range} \\ \hline
Number of Hops           & Number of Hops around the central node for extracting the subgraph                                            & Integer       & {[}1-5{]}      \\
Rollout                  & Total number of iterations for MCTS                                                                           & Integer       & {[}50-300{]}   \\
Minimum atoms            & Minimum subgraph size for the search                                                                          & Integer       & {[}1-10{]}     \\
\textit{C}               & Exploration-exploitation balance for MCTS                                                                     & Float         & {[}0.1-30.0{]} \\
Expand atoms             & Number of nodes added or removed to generate new subgraphs                                                    & Integer       & {[}1-5{]}      \\
Local radius             & Radius considered for calculating the Shapley values                                                          & Integer       & {[}1-5{]}      \\
Sample number             & Number of samples considered in each MCTS iteration                                                          & Integer       & {[}1-5{]}      \\
Maximum nodes            & \begin{tabular}[c]{@{}c@{}}Maximum size allowed in the final subgraph obtained \end{tabular} & Integer       & {[}2-40{]}     \\ \hline
\end{tabular}
}
\caption{Search range of the different input hyperparameters for SubgraphX, and a brief explanation.}
\label{table:hyperparameters}
\end{table}

Our objective is to ensure high fidelity and sparsity of the evaluated graphs. In the case of SubgraphX, the fidelity of the graph is determined through the masked ($S_{m}$) and unmasked ($S_{u}$) Shapley values, according to Equation \ref{eq:fidelity}, obtained from the original SubgraphX paper.

\begin{equation}
    Fidelity = 1 - |S_{m} - S_{u}|
    \label{eq:fidelity}
\end{equation}

Although, according to Equation \ref{eq:fidelity} we will obtain high fidelity as long as $S_{m}$ and $S_{u}$ are close to each other, obtaining high values of both is fundamental since it indicates the confidence of the trained model in the obtained subgraph ($S_{m}$) as well as in the original graph ($S_{u}$). If these values are very high, clear patterns have been found in the graphs for the given classification, achieving high confidence both in the explanation method and the model trained. Therefore, the hyperparameter tuning should seek to maximise the masked, unmasked and fidelity score for the considered dataset according to Equation \ref{eq:optimization_formula}. 

\begin{equation}
\frac{1}{3} \left( \sum_{S_{m}=1}^{n} S_{m} + \sum_{S_{u}=1}^{n} S_{u} + \sum_{S_{s}=1}^{n} S_{s} \right)
\label{eq:optimization_formula}
\end{equation}
\[
\textrm{where } n = \textrm{Number of samples}
\]

Equation \ref{eq:optimization_formula} represents the weighted sum of these three values according to the graph samples and then weights these summations. All the parameters considered for optimisation are bounded between [0,1] and give equal weight to the three parameters considered. This step is illustrated in Figure \ref{fig_architecture}: Post-hoc Explainability Optimisation. 

\subsection{Determine Most Important Subgraph}
Through the optimisation performed by AutoGOAL, we use the derived hyperparameters to obtain the most critical subgraphs of the task in question, focusing on the test data, as indicated in Figure \ref{fig_architecture}: Most Important Subgraph.

\subsection{Graph Analysis} \label{analysis}

 The analysis of the graph and the retrieved subgraph is divided into two parts: semantic analysis, which considers the terms associated with the various nodes and relationships in the graph, and structural analysis, which examines the correlations based on the different properties of the graph for both correctly and incorrectly classified graphs. Figure \ref{fig_architecture}: Graphs Analysis illustrates this step. Considering the given explainability module, SubgraphX, the masked scores give us an idea of the importance of a subgraph within the original graph for the final prediction of the model. In contrast, the unmasked scores determine the importance of this subgraph if it is absent. In this way, different conclusions can be drawn depending on the difference between the two values, as shown in Figure \ref{fig:interpretation}:

\begin{figure}
    \centering
    \includegraphics[width=0.7\linewidth]{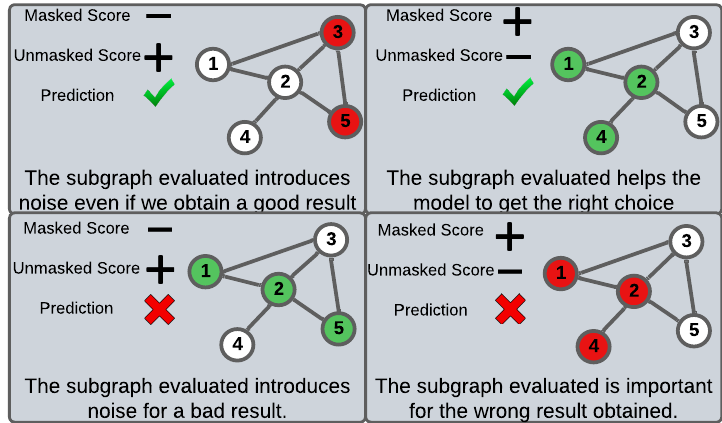}
    \caption{Interpretation of the subgraph obtained according to the classification performed. We see the positive subgraphs in green to obtain a correct prediction, while in red, we have the opposite case.}
    \label{fig:interpretation}
\end{figure}

\begin{itemize}
    \item If $S_m > S_u$, the subgraph is essential in determining the model's prediction.

    \item If $S_m < S_u$ the considered subgraph adds noise to the prediction obtained.
\end{itemize}

From a mathematical perspective, we can define two sets of disjoint and complementary subgraphs, as each subgraph will fall into the category of essential ($E$) or noisy ($N$) subgraph as shown in EquationS \ref{EN_sets} and \ref{eq:indicators}.

\begin{equation}
    \left\{
    \begin{array}{l}
        E = \{ i \in \{1,\dots,n\} : S_m^{(i)} > S_u^{(i)} \} \\[8pt]
        N = \{ i \in \{1,\dots,n\} : S_m^{(i)} < S_u^{(i)} \} \\[8pt]
        E \cup N = \{1,\dots,n\}, \quad E \cap N = \varnothing.
    \end{array}
    \right.
    \label{EN_sets}
\end{equation}

\begin{equation}
E^{(i)} = 
\begin{cases}
1, & \text{if } S_m^{(i)} > S_u^{(i)}, \\
0, & \text{otherwise},
\end{cases}
\quad
N^{(i)} = 
\begin{cases}
1, & \text{if } S_m^{(i)} < S_u^{(i)}, \\
0, & \text{otherwise}.
\end{cases}
\label{eq:indicators}
\end{equation}

However, these scores focus on the prediction obtained, not on whether the prediction is correct. If we know the veracity of the prediction, the masked/unmasked score can give much more valuable information on where the model fails:

For correct predictions, i.e. predictions in set $\mathcal{C}$: 

\begin{itemize}
    \item The subgraph with the most masked score tells us which subgraphs are essential, allowing it to get the classification right.

\begin{equation}
i_{essential} = \arg\max_{i \in \mathcal{C}} \left\{ S_m^{(i)} \cdot E^{(i)} \right\}, \quad \text{with } S_m^{(j^*)} > S_u^{(j^*)},
\label{eq:correct_learned}
\end{equation}

    \item The subgraph with the most unmasked score tells us which section of the graph introduces the most noise when figuring out the correct choice.

\begin{equation}
i_{noise} = \arg\max_{i \in \mathcal{C}} \left\{ S_u^{(i)} \cdot N^{(i)} \right\}, \quad \text{with } S_u^{(k^*)} > S_m^{(k^*)}.
\label{eq:correct_noise}
\end{equation}

\end{itemize}

For incorrect predictions, i.e. predictions in set $\mathcal{I}$:

\begin{itemize}
    \item The subgraph with the most masked scores tells us the part of the graph that implies this wrong classification.

\begin{equation}
i_{wrong} = \arg\max_{i \in \mathcal{I}} \left\{ S_m^{(i)} \cdot E^{(i)} \right\}, \quad \text{with } S_m^{(j')} > S_u^{(j')},
\label{eq:incorrect_influential}
\end{equation}

    \item The subgraph with the most unmasked scores indicates the part of the graph that is neglected by the model but may be significant in changing the model's prediction.

\begin{equation}
i_{neglected} = \arg\max_{i \in \mathcal{I}} \left\{ S_u^{(i)} \cdot N^{(i)} \right\}, \quad \text{with } S_u^{(k')} > S_m^{(k')}.
\label{eq:incorrect_neglected}
\end{equation}

\end{itemize}

\subsubsection{Semantic Analysis} \label{semantic_analysis}

\begin{figure}[t]
    \centering
    \includegraphics[width=0.9\linewidth]{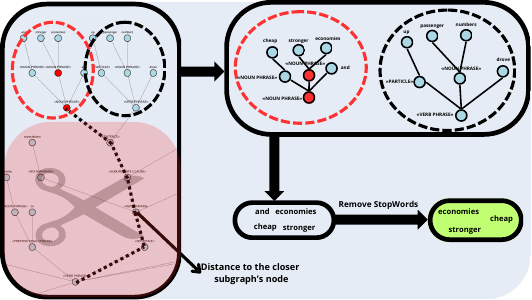}
    \caption{Methodology to obtain the most important words with semantic meaning, the minimum distance between the root node and the subgraph is obtained, dropping all the nodes with a distance below that minimum. From the obtained clusters, we keep only the one that contains the subgraph and then explore the graph to get the nodes with word labels $v_w$.}
    \label{fig:analysis_example}
\end{figure}

In addition, once the most important subgraph is determined, we can retrieve the words with semantic meaning associated with this subgraph by following the hierarchical relationships of the tree. The approach to obtain the words related to this subgraph is presented in Algorithm \ref{alg:process_hierarchical_graph} and illustrated in Figure \ref{fig:analysis_example}. In this way, we determine the set of words considered essential by the model, depending on the label considered and which words are excluded from this classification. This set of words can be subdivided according to whether the model has succeeded or failed in the given classification, thus being able to detect terms in which the model tends to fail when classifying. Hence, we have a set of entities considered important by the model for the different classifications obtained.

\subsubsection{Structural Analysis}

The semantic analysis discussed in \ref{semantic_analysis} is defined by the linguistic structure and associated with the initial data text format. However, using graphs for the final task makes it possible to perform a structural analysis of the graphs to determine which structural features are associated with correct classifications within the model. In this sense, different aspects of the graph structure can be considered to find correlations or outliers within the examples used:

\begin{itemize}
    \item General properties, such as the number of nodes and edges and different degree metrics of each graph.

    \item Community properties: in this case, different clusters are formed in each graph, as well as the connectivity and centrality of the graph.

    \item Shape properties: cycles, cliques and shapes obtained from the connection of an arbitrary number of nodes (triangles, rectangles, etc).

    \item Spectral properties: these properties are derived from the Laplacian matrix of the graph and are often used as general properties of the graph itself.
\end{itemize}

In this case, non-cyclic directed graphs are used, which can be considered hierarchical. The hierarchies are also strongly defined, so each leaf has a unique root. Specific properties of those mentioned, such as shape, spectral, and some community properties, do not make sense in this case. Still, it is worth considering them for future problems in which the graphs used do not have these characteristics.

\section{Experiments} \label{experiments}

The methodology explained in the previous section is modular, allowing the exchange of different components such as the language model used, the model for converting text into graphs, the graph neural network or the explainability module. In section \ref{configuration}, we discuss the exact configuration used in our experiments, while the evaluation results are reported in section \ref{results}.

\subsection{Configuration} \label{configuration}

To ensure the classification's explainability, we first fine-tuned the original language model on the desired tasks, saving the best model to generate the graph embeddings to train the GNN. BERT is our chosen language model for this step. To train the GNN, a GCN is used as a specific module \cite{kipf2017semisupervisedclassificationgraphconvolutional} together with an MLP as a classification head. The results obtained from the fine-tuned language model are used as labels for the GNN.

This methodology employs post-hoc methods to obtain knowledge of the original sentences. In this case, SubgraphX is the chosen module since it is designed to get the most critical subgraphs of the initial graph. Thus, it allows us to know the most important relationships for each classification obtained and maintains traceability over the initial graph, unlike other models that can obtain subgraphs with isolated nodes.

Along with SubgraphX, AutoGOAL is used as a hyperparameter optimisation framework, automatically determining the optimal input parameters for SubgraphX according to the graphs to be evaluated. This is vital to obtain significant subgraphs for the given classification (high fidelity) while reducing the subgraph size obtained (high sparsity). Once the explanations with high confidence are generated, it is possible to extract valuable information from the explanations given as a function of the modulus of explainability used; in the case of SubgraphX, the information can be derived from the Shapley values obtained according to Figure \ref{fig:interpretation}.

\subsection{Datasets}

To determine the essential words and relations, we use relatively simple sentiment analysis and topic detection datasets such as \textit{AG News} \cite{10.5555/2969239.2969312} and \textit{SST-2} \cite{socher-etal-2013-recursive}. Obtaining a good score on the task to be performed with the LLM and the GNN is crucial since the knowledge obtained from the language model is used to train the GNN, and in case of poor results, this error is propagated in successive tasks. In addition, the explainability obtained in GNNs that perform poorly may indicate which patterns of the input data it fails to represent correctly, but it can hardly provide patterns of interest for future tasks.

\subsection{Evaluation Results}  \label{results}

Table \ref{table:f1} shows the precision, recall, and F1 results obtained for the fine-tuned language model, together with the results of training the GNN with the LM classifications.

\begin{table}[]
\centering
\begin{tabular}{cccc|ccc}
\cline{2-7}
                 & \multicolumn{3}{c|}{\textbf{BERT}}       & \multicolumn{3}{c}{\textbf{GCN}}         \\ \cline{2-7} 
                 & \textit{R} & \textit{P} & \textit{$F_1$} & \textit{R} & \textit{P} & \textit{$F_1$} \\ \hline
\textbf{Ag-news} & 0.94       & 0.94       & 0.94           & 0.93       & 0.93       & 0.93           \\
\textbf{Sst-2}   & 0.92       & 0.92       & 0.92           & 0.85       & 0.85       & 0.85           \\ \hline
\end{tabular}
\caption{Results of the classifications performed in each task: on the left, we have the fine-tuned language model, and on the right, the GNN tries to emulate its behaviour. \textit{R} stands for \textit{Recall} and \textit{P} for \textit{Precision}.}
\label{table:f1}
\end{table}

Explainability is carried out on validation or test dataset samples according to their availability. 

To validate our hypothesis, we performed a general hyperparameter search on a set of data to be explained, concretely the test dataset for \textit{AG News} and the validation dataset for \textit{SST-2}, using a set of 100 samples in each case. The best hyperparameters for this set are determined according to a 24-hour experiment. After determining the optimal hyperparameters, the important subgraphs for the whole set of test in \textit{AG News} and development in \textit{SST-2} have been obtained. The results obtained according to Formula \ref{eq:optimization_formula} together with the obtained hyperparameters are shown in Table \ref{optimization_results}.

We conducted a semantic analysis that examined the important entities of the graphs obtained for each given label. In this case, the entities found in each subgraph for each category have been considered, grouping the entities by their parent phrases from the constituency tree. This information can be found in Figure \ref{fig:wordclouds_agnews}.

Correlations have been found after the structural analysis of the different graphs, as seen in the heat maps in Figure \ref{fig:Sports_ag_news}, which differ for correctly and incorrectly classified graphs.

\begin{table}[]
\centering
\begin{tabular}{ccc}
\cline{2-3}
\multirow{2}{*}{\textbf{Hyperparameter}} & \multicolumn{2}{c}{\textbf{Dataset}} \\ \cline{2-3} 
                                         & \textbf{\textit{AG News}}   & \textbf{\textit{SST-2}}   \\ \hline
Number of Hops                           & \textit{4}         & \textit{2}      \\
Rollout                                  & \textit{300}       & \textit{300}    \\
Minimum atoms                            & \textit{3}         & \textit{1}      \\
C                                        & \textit{20}        & \textit{1}      \\
Expand atoms                             & \textit{4}         & \textit{5}      \\
Local radius                             & \textit{3}         & \textit{5}      \\
Number of samples                        & \textit{5}         & \textit{5}      \\
Maximum nodes                            & \textit{2}         & \textit{2}      \\ \hline
\textbf{Optimisation Results}            & \textit{0.87}      & \textit{0.79}   \\ \hline
\end{tabular}
\caption{Best Results obtained for our optimisation with AutoGOAL.}
\label{optimization_results}
\end{table}

\section{Discussion} \label{discussion}

The method used to obtain explainability in language processing carries out several concatenated tasks so that the error of each task is propagated to the next. Nevertheless, the results obtained in the classification by the GNN are promising since the overall F1 obtained by the GNN for the initial task reaches a value of 0.87 for \textit{AG News} and 0.78 for \textit{SST-2} as indicated in Table \ref{optimization_results}.

The subgraph information obtained from SubgraphX can benefit from correct and incorrect results, as shown in Figure \ref{fig:interpretation}. Accordingly, we can determine which constituency elements of the original sentence give a wrong classification for a specific task. Based on the values obtained during the SubgraphX optimisation, this analysis is highly confident and can help diagnose our model. Note that the model's confidence comes from its knowledge and does not have to correspond to human common sense.

\begin{figure}[!htbp]
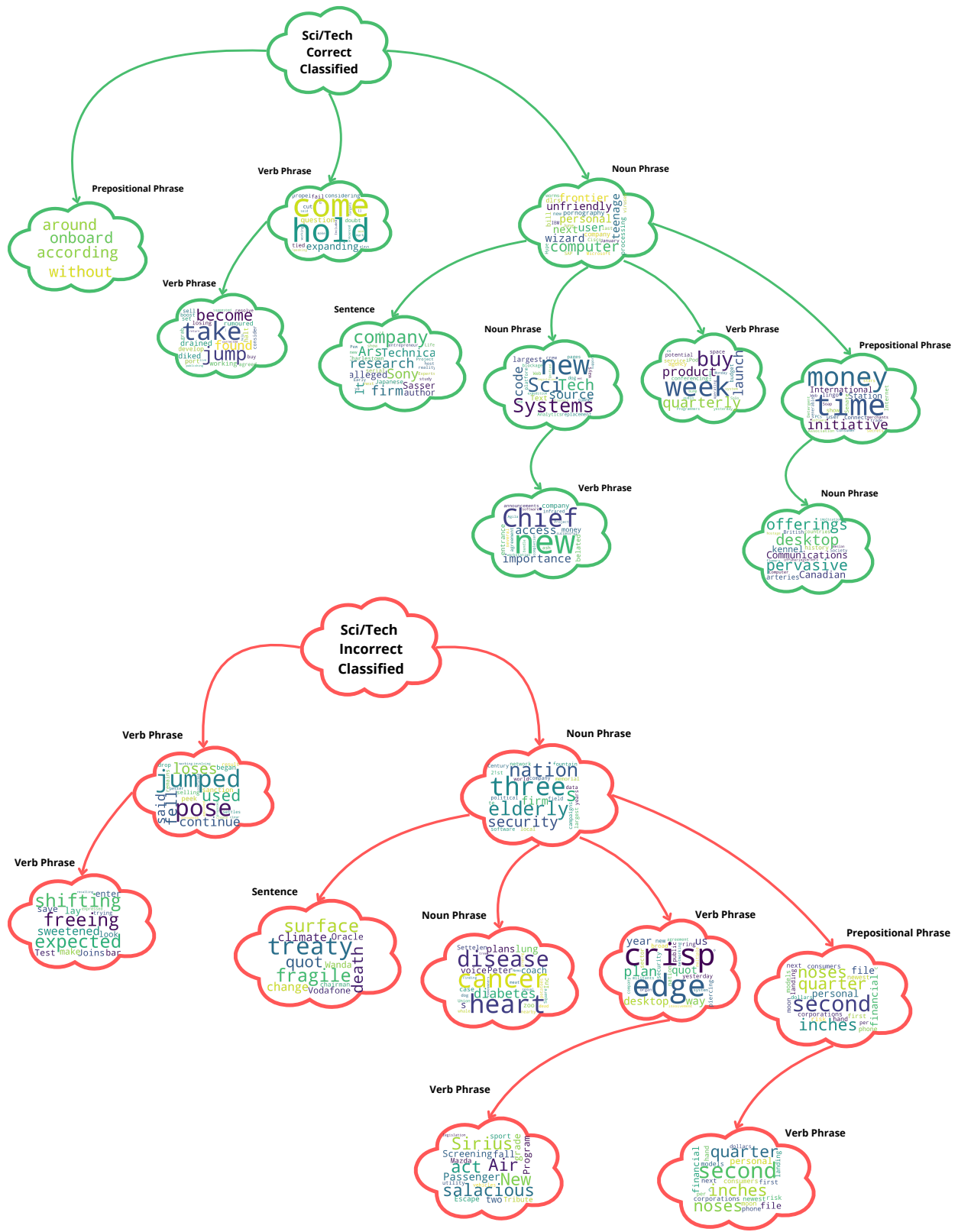

    \centering
    \begin{subfigure}[b]{\linewidth}
        \centering
        \includegraphics[width=\linewidth]{Sci-Tech_Correct.pdf}
    \end{subfigure}
    
    \begin{subfigure}[b]{\linewidth}
        \centering
        \includegraphics[width=\linewidth]{Sci-Tech_Incorrect.pdf}
    \end{subfigure}
    
    \caption{We generate word clouds for subgraphs classified as "Sci/Tech" in the \textit{AG News} Dataset. We group the obtained words by their parent phrases according to the constituency tree structure and represent their frequency as word clouds.}
    \label{fig:wordclouds_agnews}
\end{figure}

To visualise the importance of each word in the groups considered, we have used word clouds for the important subgraphs, grouping them according to the sentences from which they come in the subgraph in question, as they are constituency trees. Figure \ref{fig:wordclouds_agnews} shows the trees obtained for the ‘Sci/Tech’ class in the \textit{AG News} dataset. In this case, for correctly classified sentences, words like ‘computer’, ‘unfriendly’, or ‘IBM’ are among the words determined as important that come directly from a \textit{Noun Phrase}, according to the groups of words considered. Other words such as ‘Sci’, ‘Tech’, or ‘Code’ come from a \textit{Noun phrase} followed by another \textit{Noun phrase} according to the hierarchy of the constituency trees.

\begin{figure}[!htbp]
    \centering
    \begin{subfigure}[b]{\linewidth}
        \centering
        \includegraphics[width=\linewidth]{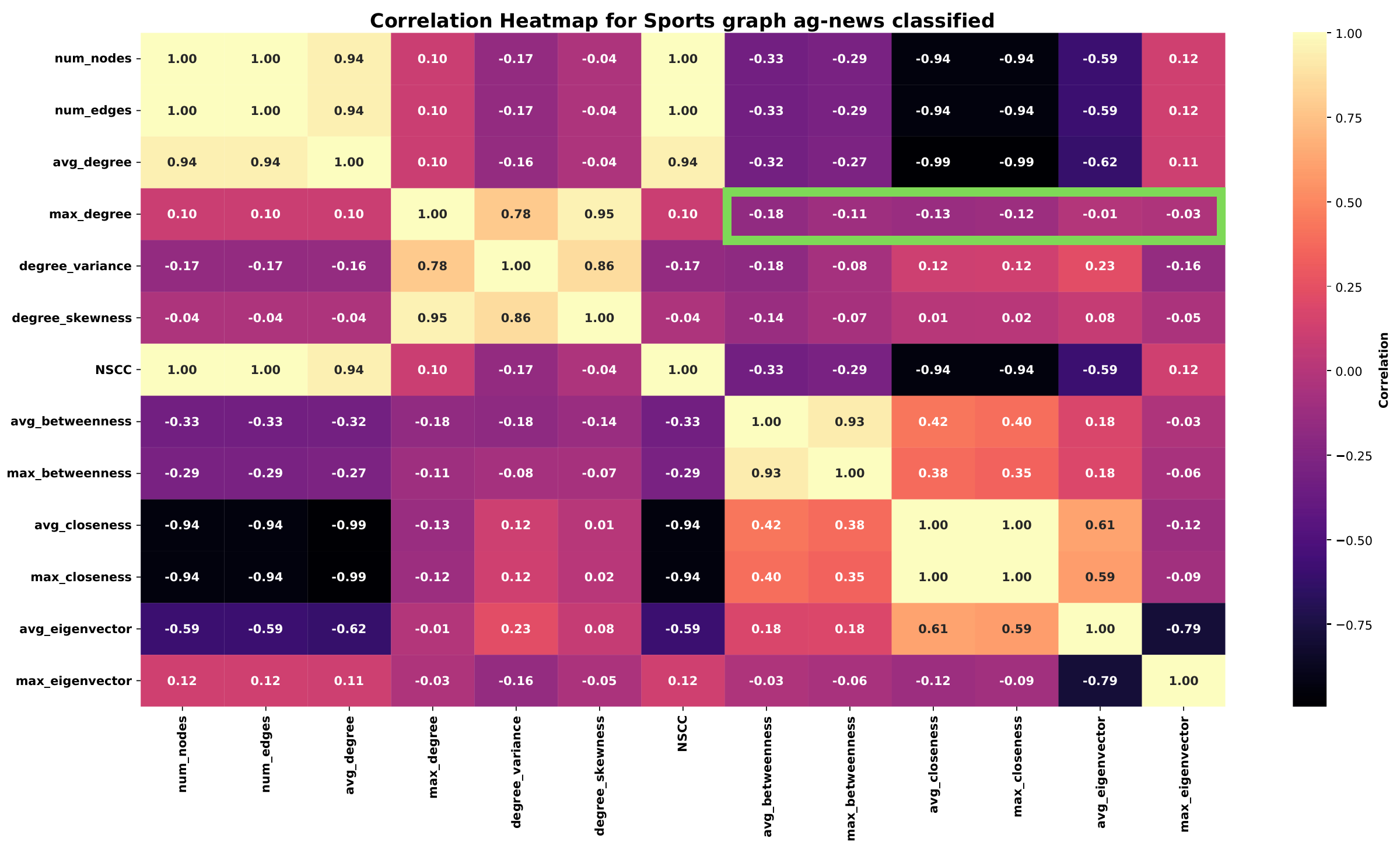}
    \end{subfigure}
    
    \begin{subfigure}[b]{\linewidth}
        \centering
        \includegraphics[width=\linewidth]{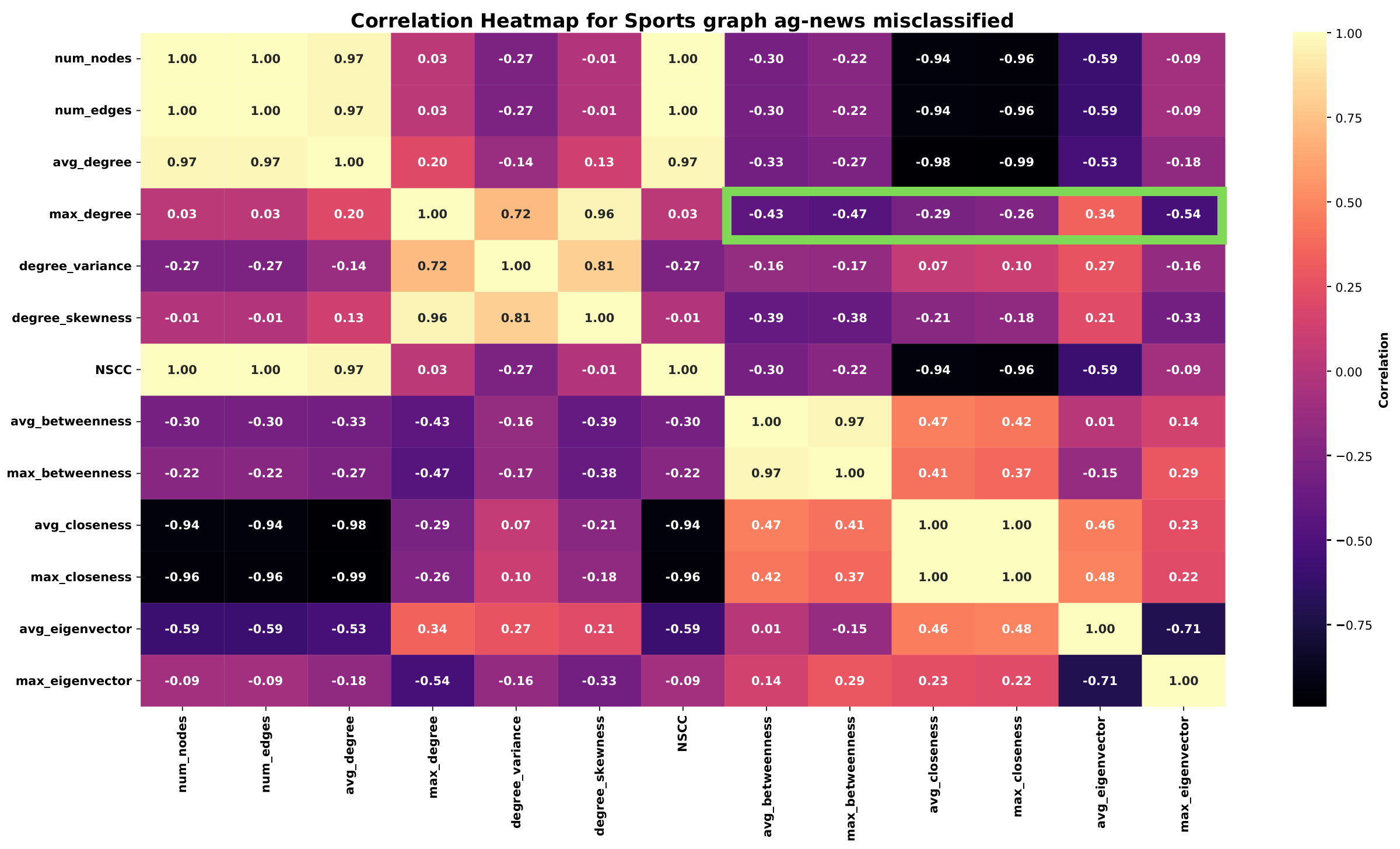}
    \end{subfigure}
    
    \caption{Structural Properties for graphs classified as "Sports" in dataset \textit{AG News}.}
    \label{fig:Sports_ag_news}
\end{figure}

Moving on to the structural analysis,  we highlight the structural properties for the ‘Sport’ category in the \textit{AG News} dataset. In this case, we found that graphs incorrectly classified in this category have an inverse correlation between the maximum degree and the average and maximum betweenness, average and maximum closeness, and maximum eigenvector. In contrast, the maximum degree positively correlates with the average eigenvector. These differences between correctly and incorrectly classified graphs indicate that misclassified graphs have a node with many relationships (high degree), while the rest of the nodes have a peripheral role (negative betweenness and closeness). Likewise, this node with many connections is very influential in the graph (positive average eigenvector). Correctly classified graphs do not show this trend, as shown in Figure \ref{fig:Sports_ag_news}. 

Semantic and structural analysis results have been obtained for each label in the datasets, including correctly and incorrectly classified graphs. This publication presents only the examples shown in Figures \ref{fig:wordclouds_agnews}, \ref{fig:Sports_ag_news}. However, the complete analysis is available on GitHub\footnote{\href{https://anonymous.4open.science/r/Language-Model-Distillation-Using-Knowledge-Graphs-for-Enhancing-Explainability-4CE7}{Github Repository With Performed Analysis}.}, where each cluster's most significant bigrams and statistical analysis are shown.

By restricting the model to a specific task, the patterns obtained from the graphs can be used to associate them with different categories according to the classifications obtained. Due to the high confidence of the generated subgraph, they are suitable for generating knowledge graphs that represent the model's knowledge. 

The methodology proposed in Figure \ref{fig:analysis_example} is adequate for hierarchical trees, as in the case of constituency trees. Still, if there are cross relations in the graph, this methodology is no longer acceptable, discouraging the combination of different types of graphs to try to obtain more information. Furthermore, determining the distance of the nodes of the explanation to the subgraph makes sense when using SubgraphX since it returns the explanation as a subgraph with all its nodes connected. This approach can give bad results if the obtained subgraph is very close to the root node, thus considering the majority of the original graph; however, these cases are easy to detect through the obtained distance. 

In this sense, the methodology used to analyse the explanation and obtain knowledge depends on several variables: the graph's structure, the explainability module used, and the objective of the task. Since the sentences are translated to data structures with traceability and use deep learning models that inherit this traceability, we can use all these variables to design explanations that the user can understand and visualise, thus better understanding the trained model.

The methodologies that directly apply explainability to LLMs require fewer steps than the proposed methodology since, in this case, numerous steps are performed in batches. However, LLMs are only used in the first stages of this methodology to perform fine-tuning and obtain embeddings, while in the methods that apply explainability directly to the LLM, successive forward and sometimes backward steps are performed, having a potentially higher computational cost. This hypothesis holds for relatively small, straightforward GNNs that can still represent the LLM's specific task.

\section{Conclusions} \label{conclussion}

By translating sentences into graphs automatically, we have performed typical LM classification tasks with GNNs, obtaining outstanding results and substantial knowledge from the generated graphs compared with the original sentences. This facilitates the diagnosis of models, the analysis of trends associated with the task to be performed, and the generation of structured knowledge for subsequent tasks.

The optimisation of the hyperparameters for the explainability module indicates that the explanations obtained from the graphs are highly confident. Thus, correctly classified results in the given tasks can be utilised in subsequent tasks. 


As far as we know, there is no similar approach in natural language processing tasks; some experiments use the distillation of LMs to GNNs with contrastive learning to align the vector space of both models and improve typical graph tasks, as is the case of LinguGKD \cite{hu2024largelanguagemodelmeets}. Other cases distil the knowledge of the language model into the GNN, using the language model to generate labels on graph nodes to subsequently do node classification with the trained GNN \cite{li2024enhancinggraphneuralnetworks}. Some have tried a different approach, distilling knowledge from GNNs to language models so that this model acquires the capabilities of GNNs \cite{mavromatis2023traingnnteachergraphaware}.

The analysis of SubgraphX highlights the distinctions between the classifications produced by the language model and the original labels of the dataset. Identifying elements deemed important and unimportant by the graph can provide insights into the significant components of the language model, especially since the GNN has been trained using embeddings from this model. Furthermore, explainability modules can be utilised to identify the key elements in the provided classification and clarify the most critical components for obtaining alternative classifications. This approach enables an examination of cases where the language model's label diverges from the actual dataset label or the GNN's classification, allowing for a detailed exploration of the original graph using specific explainability tools and visualising these discrepancies in a manner that is comprehensible to humans.

These results provide us with explainable knowledge, which, although it does not come directly from the LLM, comes from a model obtained from the distillation of that LLM, indirectly improving the explainability of the results obtained by fine-tuned LLMs in different tasks. In this aspect, we can use the explainable knowledge to have more transparent models as suggested by \cite{luo2024understandingutilizationsurveyexplainability}.

\section{Future Work} \label{future_work}

Future work should focus on the automatic population of knowledge bases and leverage these resources to enhance tasks that rely on specific knowledge, as this methodology converts unstructured knowledge into structured knowledge suitable for knowledge bases. Another important direction involves diagnosing models by identifying the patterns they frequently fail and using these insights to improve their performance. In particular, we will focus on the differences in structural properties for specific masking; the semantic analysis can also enhance the masking strategy.

Moreover, the proposed methodology can be expanded to include graphs enriched with more information than the original sentences, either through specialized knowledge graph construction, dependency parsing, or hybrid approaches. This strategy allows for the incorporation of heterogeneous graphs, which encode different types of relationships between nodes, thereby simplifying the combination of multiple data sources. 

Further research could explore how various language models, such as DeBERTa, process their knowledge differently when integrated with GNNs due to their unique internal structures. 

Lastly, exploiting the inherent structure of other explainability modules—beyond Shapley values—could be useful for gaining deeper insights into model predictions, especially when employing different fidelity metrics to capture various aspects of a model’s decision-making process.

\bibliographystyle{elsarticle-harv}

\bibliography{references}


\appendix
\clearpage

\section{Special Nodes} \label{tables}

\begin{longtable}{|>{\centering\arraybackslash}m{0.05\textwidth}|>{\centering\arraybackslash}m{0.25\textwidth}|>{\raggedright\arraybackslash}m{0.65\textwidth}|}
\caption{Terms used for the constituency special nodes with detailed explanations.}\label{tbl:constituents}\\
\hline
\textbf{Id} & \textbf{Constituency Node} & \textbf{Explanation} \\ 
\hline
\endfirsthead
\multicolumn{3}{c}%
{{\bfseries \tablename\ \thetable{} -- continued from previous page}} \\
\hline
\textbf{Id} & \textbf{Constituency Node} & \textbf{Explanation} \\ 
\hline
\endhead
\hline \multicolumn{3}{r}{{Continued on next page}} \\ \hline
\endfoot
\hline
\endlastfoot
1  & SENTENCE & Represents the entire sentence and serves as the root of the parse tree. Note that embedded clauses that form complete propositions (with their own subject--predicate structure) are also labeled as SENTENCE to maintain a consistent treatment of full propositions. \\ 
\hline
2  & NOUN PHRASE & Built around a noun (or pronoun) and functions as a subject, object, or complement. May include determiners, adjectives, and modifiers that further specify the entity or concept. \\ 
\hline
3  & VERB PHRASE & Centers on the main verb and includes any auxiliaries, objects, or complements. It represents the predicate and encapsulates the action or state expressed by the sentence. \\ 
\hline
4  & PREPOSITIONAL PHRASE & Begins with a preposition and includes its object (typically a noun phrase). Functions as an adjunct to provide additional details such as time, location, or manner. \\ 
\hline
5  & ADJECTIVE PHRASE & Centers on an adjective which may be modified by adverbs or complemented by additional elements. Functions to describe or qualify a noun either before or after it. \\ 
\hline
6  & ADVERB PHRASE & Composed of an adverb (or adverbial group) that modifies a verb, adjective, or another adverb, conveying details about time, manner, degree, or place. \\ 
\hline
7  & SUBORDINATE CLAUSE & A dependent clause introduced by a subordinating conjunction (e.g., “because,” “although”). It provides additional information (such as reason, condition, or time) and cannot stand alone as a complete sentence. \\ 
\hline
8  & PARTICLE & A function word (often non-inflecting) that attaches to a verb to form a phrasal verb or to modify the verb’s meaning. Examples include “up” in “pick up” or “off” in “take off.” \\ 
\hline
9  & INTERJECTION & A standalone exclamatory word or short utterance that expresses emotion or reaction. Typically punctuated and separate from the main grammatical structure. \\ 
\hline
10 & CONJUNCTION PHRASE & A constituent that groups elements connected by a coordinating conjunction (such as “and” or “or”). It clarifies how multiple coordinated elements function as a single unit. \\ 
\hline
11 & LIST MARKER & Identifies tokens or markers (e.g., bullets or numbers) that signal the beginning of a list item in enumeration. \\ 
\hline
12 & UNLIKE COORDINATED PHRASE & Used when coordinated elements do not share the same syntactic head or are heterogeneous. This label differentiates such structures from standard coordinated phrases. \\ 
\hline
13 & PARENTHETICAL & Represents an aside or extra information inserted into the sentence (often enclosed in commas, dashes, or parentheses) that supplements the main clause without altering its core structure. \\ 
\hline
14 & FRAGMENT & A sequence of words that conveys meaning but does not form a complete sentence (for example, lacking a subject or a predicate). Common in informal language or titles. \\ 
\hline
15 & INVERTED SENTENCE & Denotes sentences with non-standard word order (such as fronting elements for emphasis), where the canonical subject--verb order is reversed. \\ 
\hline
16 & SUBORDINATE CLAUSE QUESTION & A subordinate clause in the form of a question, typically introduced by a wh-word, that functions as a dependent clause within a larger sentence. \\ 
\hline
17 & QUESTION & A complete interrogative sentence designed to elicit information. It is parsed as an independent sentence with its own syntactic structure. \\ 
\hline
18 & WH-ADJECTIVE PHRASE & Contains a wh-word functioning adjectivally to modify a noun (e.g., “which” in “which book”), often used in questions or relative clauses. \\ 
\hline
19 & WH-ADVERB PHRASE & Composed of a wh-adverb that modifies a verb, adjective, or entire clause (e.g., “how,” “when”), typically in interrogative contexts. \\ 
\hline
20 & REDUCED RELATIVE CLAUSE & A compact form of a relative clause that omits the relative pronoun (and sometimes auxiliary verbs) yet still modifies a noun (e.g., “the man seen yesterday”). \\ 
\hline
21 & NOUN PHRASE (NO HEAD) & A noun phrase where the head is absent or implicit, often resulting from ellipsis or incomplete constructions. \\ 
\hline
23 & QUANTIFIER PHRASE & Built around a quantifier (such as numerals or determiners like “all” or “many”) that specifies quantity or degree. \\ 
\hline
24 & NOT A CONSTITUENT & Marks a span of words that does not form a valid syntactic constituent, often used in error analysis or to exclude non-groupable sequences. \\ 
\end{longtable}

\clearpage

\section{Algorithm used for Semantic Analysis} \label{algorithm}

\begin{algorithm}[H]
\caption{Processing a Hierarchical Graph to Extract Labels with Semantic Meaning.}
\label{alg:process_hierarchical_graph}


\KwIn{A hierarchical graph $G=(V,E)$, where $V={v_1, v_2, \ldots, v_n}$ is the set of nodes and $E \subseteq V \times V$ is the set of edges. Each node $v \in V$ is associated with a textual label $\mathcal{L} = {l_1, l_2, \ldots, l_n}$. Let $S \subseteq V$ be the subset of nodes inducing a subgraph of interest, $r \in V$ be a designated root node, and $W$ be a set of stopwords.}

\KwOut{A filtered set of labels $\mathcal{L^*}$ derived from ${\mathcal{L} \mid v \in S}$ after removing labels containing terms from $W$.}

\textbf{Compute shortest distances for subgraph nodes:}\; 
\ForEach{$v \in S$}{
    $\mathrm{dist}(v) \gets \text{Distance}(G, r, v)$\;
}

\textbf{Determine the minimum distance:}\; 
$d_{\min} \gets \min_{v \in S} \mathrm{dist}(v)$\;

\textbf{Prune the graph:}\; 
\ForEach{$v \in V$}{
    \If{$\text{Distance}(G, r, v) < d_{\min}$}{
        Remove $v$ from $G$\;
    }
}

\textbf{Determine clusters by the different connected components, \\ $\mathcal{C} = \{ C_1, C_2, \ldots, C_\mathcal{C} \} $:}\; 
$\mathcal{C} \gets \text{FindConnectedComponents}(G)$\;
\ForEach{$C \in \mathcal{C}$}{
    \If{$C \cap S = \emptyset$}{
        $\mathcal{C} \gets \mathcal{C} \setminus \{C\}$\;
    }
}

Let $C^*$ be the unique remaining cluster in $\mathcal{C}$\;

\textbf{Collect labels via BFS:}\; 
$\mathcal{L}_c \gets \text{BFSCollectLabels}(C^*, r)$\;

\textbf{Filter non-word labels:}\; 
$\mathcal{L}_w \gets \{\ell \in \mathcal{L}_c \mid \text{IsWord}(\ell)\}$\;

\textbf{Remove stopwords:}\; 
$\mathcal{L}^* \gets \mathcal{L}_w \setminus W$\;

\Return $\mathcal{L}^*$\;
\end{algorithm}

\end{document}